\documentclass[11pt]{article}

\usepackage{acl}

\usepackage{times}
\usepackage{latexsym}
\usepackage[T1]{fontenc}
\usepackage[utf8]{inputenc}
\usepackage{microtype}
\usepackage{inconsolata}
\usepackage{graphicx}
\usepackage{float}
\usepackage{booktabs}
\usepackage{multirow}
\usepackage{array}
\usepackage{amsmath}
\usepackage{amssymb}
\usepackage[table]{xcolor}
\usepackage{tikz}
\usepackage{pgfplots}
\usepackage{tabularx}
\pgfplotsset{compat=1.18}
\usetikzlibrary{positioning,arrows.meta}

\title{Boundary Suppression Asymmetry in Post-trained Assistants:\\
Over-expansion as a Controllability Cost}

\author{Jiarui Han\\
j22han@uwaterloo.ca}

\begin{document}
\maketitle

\begin{abstract}
Post-trained language-model assistants are often optimized to avoid under-answering, encouraging complete, helpful, cautious, and proactive responses. We ask whether this optimization creates asymmetric controllability costs: when users explicitly request narrower answers, which assistant behaviors remain suppressible, and which continue to shape the response? We study this problem as \emph{boundary-suppression asymmetry}. Prompt-side probes across multiple high-level response dimensions suggest a selective cost, concentrated around ``too-much assistant'' directions such as over-completion, extra help, and anti-underanswering.

Using controlled assistant-policy variants derived from a shared base model, we find that anti-underanswering policies are harder to pull back than the baseline under matched boundary-control evaluations, while minimal-boundary variants generally avoid this anti-side upward shift in the direct boundary-control comparisons. Mechanism-oriented probes point beyond longer default outputs, pure EOS failure, uncertainty compensation, and local continuation bias, while robustness checks preserve the main anti-over-baseline ordering under shared-system and larger-scale settings. The evidence supports a mixed planning/stopping account, where content-budget overshoot and continuation persistence jointly make boundary correction harder. Overall, post-training may create direction-specific controllability costs: some helpful assistant tendencies remain easy to invoke, yet harder to locally suppress.
\end{abstract}

\section{Introduction}

Post-trained language-model assistants are often optimized against omission: they are encouraged to be helpful, complete, cautious, and unlikely to leave important information unsaid. Instruction tuning and preference-based post-training reinforce this broad direction \citep{wei2021flan,sanh2021t0,ouyang2022instructgpt,bai2022hh,rafailov2023dpo}. This pressure is understandable: under-answering is often more visible, more frustrating, and easier to penalize than mild over-completion.

Yet the same pressure becomes a liability when users need the model to stop acting like a full assistant. Sometimes the desired output is only a sentence, a label, a decision, or no surrounding assistance at all. When the model acknowledges this boundary but still adds framing, caveats, reassurance, or adjacent help, the problem goes beyond verbosity. The user has failed to recover local control over the form and scope of the answer. In this sense, over-completion becomes a localized but concrete failure of user-side controllability.

A natural hypothesis is that boundary-control failures are a general side effect of making models more assistant-like. If post-training rewards helpfulness, caution, responsiveness, and cooperation, then many high-level assistant traits might become difficult for users to locally suppress. Our exploratory prompt-side probes suggest a narrower picture across dimensions such as clarification behavior, format-like controls, affective style, and interactional style. We use these probes to delimit the phenomenon before turning to controlled post-training comparisons. The more stable failures cluster around a smaller region: answering briefly, staying within the requested scope, avoiding extra completion, and stopping after a minimally sufficient answer. This points to a selective controllability cost: post-training may leave many assistant-like behaviors locally adjustable while hardening the tendency to do more than the user asked.

We therefore study \emph{boundary-suppression asymmetry}: why some post-trained assistant policies become asymmetrically hard to suppress under explicit boundary-narrowing instructions. To make the problem analyzable, we construct controlled policy variants from the same base model, varying only the assistant-policy target: \texttt{baseline}, \texttt{anti\_underanswer}, and \texttt{minimal\_boundary}. Because these variants share the same base model, SFT pipeline, and user-task pool, this setup reduces several broad confounds and isolates a local policy-direction contrast: whether anti-underanswering changes how easily users can suppress the policy under matched boundary-control prompts.

We make four main contributions. First, we reframe over-completion as a post-training controllability problem centered on user-side boundary suppression. Second, using controlled policy variants, we identify an anti-underanswering direction that is systematically harder to pull back than the baseline under matched boundary-control probes; second-family, shared-system, and 7B-scale checks preserve the main ordering. Third, we provide mechanism-oriented evidence for a mixed planning/stopping account. Fourth, we characterize stage dynamics: later anti-oriented SFT amplifies the asymmetry, while anti-side preference-like pullback provides the clearest reversal and minimal-side midpoint recovery remains partial. External prompt-side probes then suggest that pullback failures concentrate in a ``too-much assistant'' neighborhood rather than reflecting universal instruction-following failure.

\section{Related Work}

\subsection{Instruction Following and Post-training Conventions}

The broad background is post-training for instruction following and assistant alignment. Instruction tuning, RLHF, AI-feedback pipelines, and preference-optimization methods show that post-training can systematically reshape assistant behavior \citep{wei2021flan,sanh2021t0,ouyang2022instructgpt,bai2022hh,bai2022constitutional,rafailov2023dpo,yuan2023rrhf,hong2024orpo}. IFEval studies whether models satisfy explicit instructions under verifiable constraints \citep{zhou2023ifeval}, while Instruction Hierarchy studies how models should prioritize conflicting directives across system, developer, and user levels \citep{wallace2024instructionhierarchy}. Inverse IFEval is closest in spirit because it treats post-training conventions as locally difficult to override when users request counter-conventional behavior \citep{zhang2025inverseifeval}.

Our question is narrower: whether post-training can make a specific assistant-side boundary policy difficult for users to suppress. This shifts the object from broad convention override to boundary reclaimability, and the method from benchmark-style diagnosis to controlled policy-direction analysis: we construct matched post-training families, study an anti-underanswering direction, and decompose the resulting pullback failure behaviorally.

\subsection{Length Control, Controllability, and Behavioral Failures}

Prior work shows that aligned assistants can drift toward longer responses and that explicit length constraints remain nontrivial \citep{zhao2024longismore,yuan2024lengthconstraints}; YapBench further asks whether chatbot-style assistants talk too much and when evaluators reward excessive elaboration \citep{borisov2026yapbench}. These papers motivate our setting, while our target is the residue that remains after the user has already issued a boundary-control instruction. A model can be long by default while still being easy to pull back; our object is whether a learned policy continues to shape the response after the user tries to constrain it.

Work on controllable generation provides methods for steering outputs along desired attributes \citep{dathathri2020pplm,krause2021gedi,yang2021fudge,li2021prefix,dong2023steerlm,zhang2022ctgsurvey}, and sequential post-training work studies how later stages preserve, overwrite, or interfere with earlier signals \citep{zhou2023lima,prakash2024finetuning,fernando2024forgetting}. We use these literatures as diagnostic background rather than proposing a new steering algorithm. Our focus is a behavioral failure mode of post-trained assistants: the learned tendency to do more than the user asked, and whether that tendency remains active when the user explicitly tries to suppress it.

\section{Problem Setup and Experimental Framework}
\label{sec:setup}

\subsection{Design Goal}

The goal is to turn a broad interactional observation into a controlled post-training comparison. A model may over-expand because of its base distribution, chat formatting, assistant-role framing, prompt wording, or a learned assistant-policy direction. Prompt-side and checkpoint-level comparisons can reveal candidate failures, but cannot identify which learned direction makes boundary pullback difficult. We therefore fix the base model, training pipeline, and task manifold, and vary only the assistant-policy specification and corresponding supervision.

\subsection{Controlled Assistant-Policy Families}

We construct three policy variants from the same base model. The variants share the same task pool and SFT procedure, but differ in the assistant-policy specification used to write the target responses:
\begin{itemize}
    \item \textbf{\texttt{baseline}}: answer the current question directly and proportionately, without explicit pressure toward either expansion or minimization;
    \item \textbf{\texttt{anti\_underanswer}}: avoid giving too little by adding useful context, completion, or adjacent help when a short answer would be insufficient;
    \item \textbf{\texttt{minimal\_boundary}}: respect the current user boundary and provide only the minimum sufficient answer.
\end{itemize}

The data are organized as triplets over the same \texttt{user\_text}, keeping the task fixed while varying assistant-side supervision. The evaluations compare how these learned response policies behave under the same boundary-control requests.

\subsection{Evaluation Logic: Suppressibility Rather Than Default Length}

The central evaluation target is \emph{suppressibility}: whether a learned response policy remains active after the user explicitly asks for a narrower answer. A model may be longer by default but still easy to pull back. The relevant question is therefore not average length alone, but whether boundary-narrowing instructions reduce the differences between policy variants.

We evaluate this with held-out prompt modes that instantiate three nearby forms of user-side boundary control:
\begin{itemize}
    \item \texttt{avoid\_underanswer}, which asks the model to give a sufficiently informative answer without compensating through rambling or unrelated scope expansion;
    \item \texttt{scope\_minimal\_sufficient}, which asks the model to cover the current request without expanding beyond it;
    \item \texttt{do\_not\_add\_unasked\_help}, which asks the model not to widen the task with adjacent assistance.
\end{itemize}

A pure default-length shift would predict that these boundary-control prompts bring the variants closer together. Persistent separation under the same boundary-control requests indicates a controllability difference: the learned policy direction continues to shape the response even when the user asks to constrain it.

\subsection{Experimental Setup}

The main controlled-family experiments use \texttt{Qwen2.5-1.5B} with a shared parameter-efficient SFT pipeline. Inputs are formatted as \texttt{system + user + assistant} message sequences, and the causal language-modeling loss is applied only to assistant tokens. All three variants use the same optimization recipe and task pool; the manipulated signal is the assistant-policy direction expressed through the training context and target responses.

The core Chinese family is built from 12 canonical source cases, expanded with 5 wrappers and 4 body-builder variants into 240 triplets. This compact manifold is intentional: it enables repeated comparisons over matched interaction structures, with broad task coverage left outside the main controlled test. The design sacrifices deployment breadth in order to make the policy-direction contrast local and repeated.

Held-out probes are separated from training data. The main boundary-control evaluation uses a 50-item Chinese held-out set; the Qwen forced-prefix probe uses 100 continuations after minimum-sufficient prefixes; and the planning-vs.-stopping annotation covers 80 responses. Later preference-like stages use pairwise data for amplification and pullback tests, with filtering rules described in the stage section and appendix. We also run a compact English \texttt{SmolLM2-1.7B-Instruct} replication and shared-system checks on \texttt{Qwen2.5-1.5B} and \texttt{Qwen2.5-7B-Instruct}; these test whether the direct ordering survives outside the primary Chinese Qwen setting, under a common system prompt, and at larger scale. The detailed mechanism and stage analyses remain concentrated in the primary Qwen family.

Appendix~\ref{app:evidence-tiers} separates claim-bearing evidence from partial and exploratory lines used to bound the interpretation.

\section{Main Phenomenon: \texttt{anti\_underanswer} is Harder to Pull Back}
\label{sec:phenomenon}

\begin{table}[t]
\centering
\small
\resizebox{\columnwidth}{!}{%
\begin{tabular}{lcccc}
\toprule
Model family & avoid UA & scope min & no extra help & FP mean \\
\midrule
\texttt{baseline} & 23.44 & 22.16 & 24.10 & 2.54 \\
\texttt{anti\_underanswer} & 33.78 & 40.38 & 28.64 & 4.08 \\
\texttt{minimal\_boundary} & 28.74 & 27.28 & 26.96 & 2.51 \\
\midrule
$\Delta$ anti-base & +10.34 & +18.22 & +4.54 & +1.54 \\
95\% CI ($\Delta$) & [1.78, 19.84] & [7.96, 31.68] & [-0.48, 9.78] & [-0.01, 3.66] \\
\bottomrule
\end{tabular}
}
\caption{Suppressibility gaps in the main controlled family. The first three columns report mean Chinese-character response length on a 50-item held-out set; FP reports mean continuation characters over 100 forced-prefix continuations. Final rows give anti-minus-baseline gaps and paired-bootstrap 95\% CIs.}
\label{tab:main-family-results}
\end{table}

Table~\ref{tab:main-family-results} summarizes the main controlled comparison. The boundary-control evaluations use a 50-item Chinese held-out set: the user asks the model to give a sufficiently informative answer without rambling or unrelated expansion, stay within the current request, or avoid unasked help. Under these conditions, \texttt{anti\_underanswer} remains more expansive than \texttt{baseline}. The \texttt{avoid\_underanswer} condition is especially diagnostic because it permits a reasonably fuller answer while discouraging loose expansion; a large anti-side gap therefore suggests overshoot from the anti-underanswering policy, not mere compliance with a short-answer instruction. The paired gaps are clearest under \texttt{avoid\_underanswer} and \texttt{scope\_minimal\_sufficient}; \texttt{do\_not\_add\_unasked\_help} remains positive but weaker.

The forced-prefix probe is more modest overall, but it contributes a late-continuation trace: \texttt{anti\_underanswer} stops immediately in only 48/100 cases, compared with 62/100 for \texttt{baseline} and 63/100 for \texttt{minimal\_boundary}, and the \texttt{scope\_minimal\_sufficient} slice shows the clearest continuation gap. Section~\ref{sec:mechanism} asks whether this residual expansion reflects content-budget overshoot, continuation past a sufficient answer, or both.

\section{Mechanism: From Shallow Explanations to Mixed Planning/Stopping}
\label{sec:mechanism}

\subsection{Shallow Accounts Are Insufficient}

We first test whether the suppressibility gap can be reduced to shallow generation-level accounts: failure to emit EOS, uncertainty-driven hedging, or a local preference to continue. The EOS-tail experiments compare prefix-only supervision, prefix-plus-true-EOS supervision, and prefix-plus-tailspan supervision. Adding the true EOS token helps, but supervising the tail span helps more, suggesting that stopping structure matters while a single-token EOS account is too narrow.

We then probe uncertainty directly from generation-time logits. The results do not support a simple uncertainty-compensation story: long \texttt{anti\_underanswer} responses are often \emph{more} confident, not less. Likewise, the branchpoint probe does not yield a strong enough signal to reduce the phenomenon to a local one-step continuation preference. Together with the quantitative audit in Appendix~\ref{app:mechanism-audit}, these probes narrow the mechanism space before we turn to the planning/stopping decomposition.

Table~\ref{tab:mechanism-probes} summarizes the resulting narrowing chain.

\begin{table*}[t]
\centering
\small
\begin{tabular}{p{2.2cm}p{2.9cm}p{5.0cm}p{3.7cm}}
\toprule
Hypothesis & Probe & Key result & Interpretation \\
\midrule
Pure EOS / stopping & prefix50, true EOS, tail span & EOS helps; tail span helps more & stopping matters, but EOS alone is insufficient \\
Pure uncertainty & entropy, margin, first-token stats & long anti responses are not more uncertain & does not support simple uncertainty compensation \\
Pure local continue bias & branchpoint probe & no strong enough local suffix-preference account & too weak to explain retained results alone \\
Pure wording compression & compression vs.\ pruning & compression often fails; pruning helps more & points to content budgeting, not only phrasing \\
No late persistence & forced-prefix continuation, immediate-stop counts & anti stops immediately less often; scope-minimal slice shows clearer continuation & over-expansion is not only early budgeting \\
\bottomrule
\end{tabular}
\caption{Mechanism-probe summary. The probes narrow the explanation away from shallow single-factor accounts and toward a mixed content-budgeting and continuation account.}
\label{tab:mechanism-probes}
\end{table*}

Taken together, the probes keep stopping in the explanation, but only as one part of a broader behavioral pattern: excess response mass is organized through both content budgeting and continuation persistence.

\subsection{Evidence for Both Content Budgeting and Late Continuation}

The next probes separate two ways in which over-expansion can arise. A response may include too much content before it reaches a sufficient answer, or it may reach a sufficient answer and then continue. Compression/pruning targets the first possibility; forced-prefix continuation targets the second.

For \texttt{anti\_underanswer}, \texttt{same\_information\_compression} often fails to recover a narrow answer, while \texttt{true\_pruning} helps more. This points to content selection beyond wording: the problem lies in what content the model chooses to include, not only in how efficiently it phrases that content.

Forced-prefix continuation tests the late side directly. In this setting, generation begins after a minimum-sufficient prefix, where stopping would already be a reasonable response. The overall Qwen continuation gap is modest, but \texttt{anti\_underanswer} stops immediately less often than the other variants and shows a clearer continuation gap under the \texttt{scope\_minimal\_sufficient} forced-prefix slice. This suggests that continuation pressure can remain after a sufficient answer has been reached, so early content budgeting explains only part of the effect.

\subsection{Planning/Stopping Annotation and Bundled Interpretation}

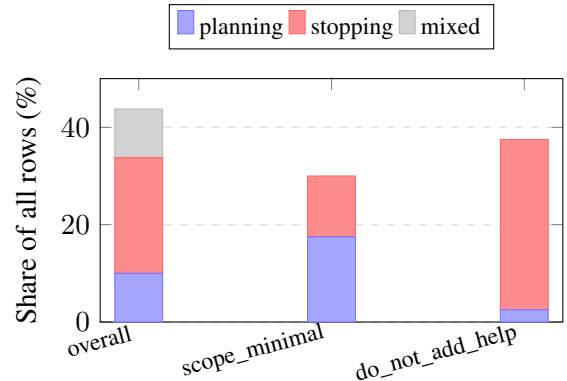
\begin{figure}[t]
\centering
\begin{tikzpicture}
\begin{axis}[
    width=\columnwidth,
    height=4.8cm,
    ybar stacked,
    bar width=18pt,
    ymin=0,
    ymax=50,
    ylabel={Share of all rows (\%)},
    symbolic x coords={overall,scope\_minimal,do\_not\_add\_help},
    xtick=data,
    xticklabel style={rotate=15,anchor=east,font=\small, yshift=-3pt},
    legend style={at={(0.5,1.12)},anchor=south,legend columns=3,font=\small},
    ymajorgrids=true,
    grid style={dashed,gray!30}
]
\addplot+[fill=blue!35,draw=blue!55] coordinates {(overall,10.0) (scope\_minimal,17.5) (do\_not\_add\_help,2.5)};
\addplot+[fill=red!45,draw=red!60] coordinates {(overall,23.75) (scope\_minimal,12.5) (do\_not\_add\_help,35.0)};
\addplot+[fill=gray!35,draw=gray!55] coordinates {(overall,10.0) (scope\_minimal,0.0) (do\_not\_add\_help,0.0)};
\legend{planning,stopping,mixed}
\end{axis}
\end{tikzpicture}
\caption{Planning-vs.-stopping failure shares over 80 annotated Qwen responses. Bars report shares of all annotated rows, so omitted mass corresponds to none/other outcomes rather than hidden normalized failure composition. Stopping failures dominate overall; \texttt{scope\_minimal\_sufficient} is more planning-heavy, while \texttt{do\_not\_add\_unasked\_help} is more stopping-heavy. Raw percentages are reported in Appendix~\ref{app:planning-stopping}.}
\label{fig:planning-stopping}
\end{figure}

The planning-vs.-stopping annotation decomposes the remaining expansion into content-level failure modes. We mark planning failures when the response expands the content budget before reaching a sufficient answer, and stopping failures when it continues after such an answer is already available. Figure~\ref{fig:planning-stopping} reports failure-type shares over all annotated responses, not a 100\% composition of only failure cases. Both failure types occur. Stopping failures dominate overall, but the mixture depends on the boundary mode: \texttt{scope\_minimal\_sufficient} is more planning-heavy, whereas \texttt{do\_not\_add\_unasked\_help} is more stopping-heavy.

The annotation gives the length gaps a content-level interpretation. The extra text is often tied to a specific boundary error: the response budgets too much content before the answer is complete, or it continues after a minimum-sufficient answer has already been delivered.

\begin{table}[t]
\centering
\small
\resizebox{\columnwidth}{!}{%
\begin{tabular}{lccc}
\toprule
Probe / pullback mode & baseline & anti & minimal \\
\midrule
\multicolumn{4}{l}{\textbf{Asymmetric controllability}} \\
\texttt{ask\_minimal} & 26.00 & 87.17 & 28.17 \\
\texttt{ask\_answer\_first} & 25.83 & 76.67 & 33.50 \\
\texttt{ask\_no\_extra\_help} & 25.83 & 49.83 & 43.33 \\
Negative-side mean & 25.89 & 71.22 & 35.00 \\
\midrule
\multicolumn{4}{l}{\textbf{Bundled generalization}} \\
\texttt{ask\_minimal} & 33.17 & 55.50 & 26.17 \\
\texttt{ask\_answer\_first} & 24.50 & 56.67 & 25.67 \\
\texttt{ask\_no\_extra\_help} & 24.00 & 58.67 & 24.67 \\
\texttt{ask\_no\_caution} & 24.00 & 65.17 & 23.50 \\
Negative-side mean & 26.42 & 59.00 & 25.00 \\
\bottomrule
\end{tabular}
}
\caption{Auxiliary bundled-control probes in the Qwen main family. Entries are mean Chinese-character lengths under pullback prompts; higher values indicate residual expansion under nearby boundary controls. The negative-side mean averages the listed pullback modes within each probe.}
\label{tab:bundle-probes}
\end{table}

Two additional held-out probe packs test whether this mixed planning/stopping picture is still too narrow. The asymmetric-controllability matrix asks whether nearby prompt toggles can independently pull the response along scope, answer-first, and extra-help dimensions. Table~\ref{tab:bundle-probes} shows that they do not fully recover the anti row: its negative-side mean remains much higher than baseline (71.2 vs.\ 25.9), while \texttt{minimal\_boundary} generally stays closer to the narrow side than \texttt{anti\_underanswer}, with some spillover on the help axis.

The bundled-generalization pack widens the neighborhood by adding caution-sensitive and one-extra-layer variants. The same pattern remains: \texttt{anti\_underanswer} stays elevated across the pullback neighborhood, with negative-side means from 55.5 to 65.2, while baseline and minimal remain near 23--33. Together, these results point away from a single length knob. The effect involves content selection and continuation pressure, above word count or a final-token stopping defect. The anti-underanswering direction changes both what content is included and how strongly the response continues after a sufficient answer; expansion pressure also generalizes across nearby controls over scope, unasked help, interactional continuation, and caution. We therefore describe the phenomenon as a behavior-level bundled assistant-policy bias.

\section{Stage Structure: Amplification and Asymmetric Reversal}
\label{sec:stages}

The mechanism results show that boundary-suppression asymmetry involves more than a simple length shift: both content budgeting and continuation pressure matter. We next use later post-training stages as interventions on this behavioral asymmetry, testing amplification, pullback, and asymmetric reversibility. The resulting pattern is asymmetric: later anti-oriented SFT shows the clearest amplification relative to the direct baseline reference, while preference-like pullback contracts the anti side more clearly than it recovers a stable midpoint from the minimal side.

\subsection{Later Anti-oriented SFT Amplifies the Effect}

A later anti-oriented SFT stage can amplify the same boundary-control cost after a baseline assistant policy has already been installed. In the \texttt{baseline\_then\_anti} line, boundary-control responses become longer, forced-prefix continuation rises sharply, and planning/stopping failures worsen. This suggests that the anti-underanswering direction can be further strengthened after the initial controlled-family split by a later stage that again rewards avoiding under-answering.

\subsection{Preference-like Same-direction Amplification Is Weaker}

The same-direction preference-like line provides weaker amplification. The \texttt{baseline\_then\_preference\_expand} runs move in the same broad direction, but the effect is smaller and non-monotonic. The strongest usable preference-stage run moves in the expected direction, while larger stable runs remain directionally consistent without further strengthening the effect. We therefore treat this as supportive evidence that preference-like training can push the same region, while leaving the primary source of the phenomenon open.

\subsection{Asymmetric Reversal: Stronger on the Anti Side}

The clearest reversal appears when preference-like pullback is applied to the anti side. In the \texttt{anti $\rightarrow$ preference\_minimal} line, boundary-control responses contract broadly, bundled-generalization behavior flattens, and forced-prefix continuation falls close to zero.

The corresponding minimal-side pullback is weaker. The \texttt{minimal $\rightarrow$ preference\_baseline} line produces only a partial midpoint recovery, and very small sweeps tend to collapse rather than improve it. Reversal capacity therefore appears asymmetric in this setting: pulling back the anti side is easier than recovering a stable midpoint from the minimal side.

\begin{table}[t]
\centering
\small
\resizebox{\columnwidth}{!}{%
\begin{tabular}{lccccccc}
\toprule
Line & neutral & avoid UA & $\Delta$ & scope min & $\Delta$ & FP & $\Delta$ \\
\midrule
\texttt{baseline\_then\_anti} & 83.00 & 92.08 & +68.64 & 97.42 & +75.26 & 55.83 & +53.29 \\
\texttt{baseline\_then\_pref\_expand} & 26.92 & 51.08 & +27.64 & 47.17 & +25.01 & 28.00 & +25.46 \\
\texttt{anti$\rightarrow$pref\_minimal} & 12.00 & 14.25 & -19.53 & 13.00 & -27.38 & 1.00 & -3.08 \\
\texttt{minimal$\rightarrow$pref\_baseline} & 14.58 & 34.58 & +5.84 & 21.67 & -5.61 & 8.42 & +5.91 \\
\bottomrule
\end{tabular}
}
\caption{Stage-split amplification and reversal results. The neutral, avoid-UA, and scope-min columns report mean Chinese-character response length; FP reports mean forced-prefix continuation characters. Delta columns compare each line with its direct-family source where available: the two baseline-started lines use the direct baseline row, the anti-pullback line uses the direct anti row, and the minimal-midpoint line uses the direct minimal row. Stage rows have 12 boundary-control rows per mode and 12 forced-prefix rows; direct-family reference rows have 50 boundary-control rows per mode and 100 forced-prefix rows.}
\label{tab:stage-and-reversal}
\end{table}

The stage results form an asymmetric pattern. Across these stage interventions, the clearest amplification comes from later anti-oriented SFT, while the clearest reversal comes from anti-side preference-like pullback. The minimal-side midpoint line is informative but weaker and more brittle. Appendix~\ref{app:stage-delta-uncertainty} reports independent-bootstrap uncertainty for the main stage deltas.

\section{Robustness and Bounded Generality}
\label{sec:external}

\subsection{Second-family Replication}

We use a compact English \texttt{SmolLM2-1.7B-Instruct} family to test whether the main ordering survives outside the primary Chinese Qwen setting. Table~\ref{tab:smollm2-transport} shows a directional replication: in the direct family, \texttt{anti\_underanswer} remains the most expansive row under boundary-control modes and forced-prefix continuation, while \texttt{minimal\_boundary} generally avoids the anti-side upward shift.

\begin{table}[tb]
\centering
\footnotesize
\setlength{\tabcolsep}{3pt}
\resizebox{\columnwidth}{!}{%
\begin{tabular}{lcccc}
\toprule
Line & avoid\_ua & scope\_min & no\_help & FP \\
\midrule
\texttt{baseline} & 229.2 & 137.7 & 174.0 & 80.3 \\
\texttt{anti\_underanswer} & 312.5 & 225.7 & 264.5 & 133.0 \\
\texttt{minimal\_boundary} & 248.8 & 154.0 & 200.2 & 97.6 \\
\midrule
\texttt{baseline$\rightarrow$anti} & 229.8 & 138.8 & 154.0 & 9.4 \\
\texttt{anti$\rightarrow$pref\_minimal} & 269.2 & 165.3 & 207.3 & 15.5 \\
\bottomrule
\end{tabular}
}
\caption{Second-family transport summary on \texttt{SmolLM2-1.7B-Instruct}. Entries are mean character lengths; FP reports forced-prefix continuation length. Direct-family rows replicate the main ordering; follow-up rows summarize partial stage transport.}
\label{tab:smollm2-transport}
\end{table}

The second-family mechanism checks point in the same direction across forced-prefix continuation, compression/pruning, and planning/stopping annotation. On SmolLM2, \texttt{anti\_underanswer} continues more after a minimum-sufficient prefix (131.5 vs.\ 72.7 mean continuation), remains harder to shorten through same-information compression (337.1 $\rightarrow$ 251.7, compared with 278.0 $\rightarrow$ 172.9 for baseline), and receives higher planning/stopping scores in a 48-row annotation pass, with 24 labeled rows per model row.

The direct ordering and mechanism checks transfer more consistently than the stage geometry. The Qwen-style later anti-oriented SFT amplification transfers less consistently, while preference-like pullback remains meaningful but incomplete, so the second-family evidence primarily supports the direct-family ordering and mechanism echoes.

\subsection{Shared-system and Larger-scale Checks}

We also test whether the direct ordering survives when response-style cues are reduced and model scale increases. In shared-system checks, the three variants are trained under a common system prompt, so the family distinction is carried by the supervised response behavior. The anti-over-baseline ordering remains in both Qwen1.5B and Qwen7B: boundary-control means are 49.75/93.30/31.55 for baseline/anti/minimal on Qwen1.5B, and 33.05/81.67/27.34 on Qwen7B. This reduces the concern that the main-family result is only an artifact of policy-specific system wording.

The 7B run also gives a stronger late-continuation trace. After a minimum-sufficient prefix, \texttt{anti\_underanswer} continues much longer than baseline (52.90 vs. 11.40) and stops immediately in only 3/100 cases, compared with 49/100 for \texttt{baseline} and 71/100 for \texttt{minimal\_boundary}. Some 7B non-scope bundled and asymmetric-control probes are axis-sensitive, so the 7B evidence is strongest for the anti-over-baseline ordering and late-continuation trace. These checks support the direct boundary-suppression and late-continuation claims beyond the primary run, while the stage-specific conclusions remain tied to the main Qwen experiments. Full summaries appear in Appendix~\ref{app:shared-system-7b}.

\subsection{External Cluster Beyond the Original Axis}

To bound generality, we run a six-model prompt-side benchmark over three nearby assistant-policy directions: \texttt{caveat/no\_extra\_caveat}, \texttt{next\_step/no\_next\_step}, and \texttt{moralizing/plain\_task\_frame}. These axes serve as clustering probes, testing whether pullback failures appear around neighboring ``too-much assistant'' tendencies.

Figure~\ref{fig:external-cluster} supports a clustered reading of the external evidence. The scores are raw annotation scores rather than normalized prevalence estimates, so they should be read together with the raw counts in Appendix~\ref{app:external-cluster-raw}. \texttt{caveat/completeness} is the most stable external prompt-side axis: it appears across models and is the axis for which we obtain a partial controlled-family extension. \texttt{next\_step/proactivity} and \texttt{moralizing} add cluster evidence, but with stronger model sensitivity. This neighborhood aligns with the controlled-family results: persistent pullback failures concentrate around boundary-facing ``too-much assistant'' behavior. The external benchmark broadens the evidence base while leaving the causal interpretation anchored in the controlled-family experiments.

\begin{figure}[tb]
\centering
\footnotesize
\setlength{\tabcolsep}{4.5pt}
\renewcommand{\arraystretch}{1.08}
\begin{tabular}{lccc}
\toprule
Model & \texttt{caveat} & \texttt{next\_step} & \texttt{moralizing} \\
\midrule
\texttt{qwen} & \cellcolor{orange!35}6 & \cellcolor{yellow!25}3 & \cellcolor{orange!30}5 \\
\texttt{ds\_v4} & \cellcolor{orange!40}7 & \cellcolor{red!45}9 & \cellcolor{orange!25}4 \\
\texttt{gpt5mini} & \cellcolor{red!50}10 & \cellcolor{orange!30}5 & \cellcolor{orange!30}5 \\
\texttt{ds7b} & \cellcolor{red!70}15 & \cellcolor{red!85}19 & \cellcolor{red!50}10 \\
\texttt{gpt5\_1} & \cellcolor{orange!25}4 & \cellcolor{yellow!20}2 & \cellcolor{orange!25}4 \\
\texttt{claude} & \cellcolor{orange!35}6 & \cellcolor{yellow!25}3 & \cellcolor{yellow!20}2 \\
\bottomrule
\end{tabular}
\caption{External cluster benchmark. Scores are raw negative/pullback-mode counts computed as $\text{partial}+2\times\text{disobey}$; higher values indicate more persistent residual tendency within the benchmark. Because axes have different item counts, the figure is an audit heatmap rather than a normalized cross-axis prevalence estimate. Raw counts and full model names are reported in Appendix~\ref{app:external-cluster-raw}.}
\label{fig:external-cluster}
\end{figure}

Additional checks reinforce this clustered reading: a small SmolLM2 discrimination check finds boundary pullback harder than formatting-only or explanation-only pullback, and a partial \texttt{caveat} controlled-family extension shows that a clear external prompt-side axis has some controlled-training analogue. Contrast axes such as \texttt{emoji\_only} and later \texttt{clarify}/\texttt{stance} diagnostics, summarized in Appendix~\ref{app:negative-controls}, delimit where the cluster does and does not appear. The external evidence therefore supports a bounded but substantive claim: pullback failures concentrate in a ``too-much assistant'' neighborhood.

\section{Discussion}

The evidence points to a structured controllability cost: a bounded ``too-much assistant'' region involving over-completion, extra help, continuation, and related completeness-oriented tendencies. This reading is narrower than generic obedience failure and broader than a single length-axis effect.

One interpretation is that the main direction lies close to a pre-existing continuation tendency in the base model, which later post-training can reweight rather than invent from scratch. A compatible interpretation is that this region bundles several desirable assistant traits---completeness, responsibility, helpfulness, caution, and low omission risk---so a user's request to keep the answer narrow pushes against a broader assistant-policy bundle. The external evidence supports a clustered boundary-recovery story rather than a universal obedience-failure account.

\section{Conclusion}

This paper studies whether assistant policies optimized away from under-answering become harder for users to locally suppress. Using controlled families, mechanism probes, stage interventions, and bounded external checks, we find that \texttt{anti\_underanswer} is harder to pull back than baseline under matched boundary-control requests, while minimal-boundary variants generally avoid the anti-side upward shift in the direct boundary-control comparisons.

The evidence supports a mixed planning/stopping account: anti-underanswering changes both what content is included and whether the response continues after a sufficient answer. Later-stage experiments show that this cost can be amplified and partially reversed, with different reversal strength across directions. More broadly, post-training can reshape not only answer style, but also how easily users recover local control over response form and scope.

\section*{Limitations}

The controlled-family experiments use a deliberately compact manifold built from canonical cases, wrappers, and policy-specific target responses. This design is useful for isolating assistant-policy direction under matched training and evaluation conditions, while broader deployment-distribution coverage remains outside its scope. The results should therefore be read as controlled evidence about a specific boundary-suppression mechanism, with no claim to cover all user requests or all assistant behaviors.

One remaining confound is target length. The three policy targets are not length-balanced: \texttt{anti\_underanswer} targets are usually longer than \texttt{baseline} targets, and this gap is larger than the gap between \texttt{baseline} and \texttt{minimal\_boundary}. The trained model may therefore inherit some asymmetric length prior from the targets.

A fully length-matched dataset would require rewriting all three target families. Still, the diagnostics suggest that length transfer alone gives an incomplete explanation. Under boundary-control evaluation, the learned anti-minus-baseline gap is larger and more stable than the baseline-minus-minimal gap, and early-stop / forced-prefix probes show anti-side continuation even when shorter continuations are available. We therefore avoid claiming a length-matched causal estimate. A fully length-balanced rewrite would be a useful ablation, but it could also introduce new target-quality confounds.

The shared-system checks address the most direct system-wording concern. A fully cue- and length-balanced rewrite would still be useful for separating target style, target length, and policy semantics more completely.

The mechanism claims are behavioral and stage-structured, leaving circuit-level analysis for future work. We identify a stable mixed planning/stopping pattern: the anti-underanswering direction affects both what content is included and whether the model continues after a sufficient answer. The current evidence identifies the behavioral pattern, while leaving open its internal implementation and how similarly that implementation appears across model families.

The external-validity evidence has an intentionally bounded scope. The controlled-family direction, \texttt{anti\_underanswer}, and the clearest external prompt-side member, \texttt{caveat/completeness}, may be related without being isomorphic. The controlled-family experiments test which directions can be cleanly amplified or partially reversed under compact post-training, while the external benchmark tests how residual pullback failures are distributed across already post-trained models. Additional high-level directions did not yield clean retained axes, so the paper's generality claim remains clustered. The evidence supports a bounded ``too-much assistant'' region, not a general failure of instruction following.

\bibliography{directional_control_research_refs}
\appendix

\section{Mechanism-probe Details}
\label{app:mechanism-audit}

\noindent Table~\ref{tab:mechanism-evidence-boundary} summarizes how the mechanism probes support the behavioral interpretation. The main evidence supports a mixed content-budgeting and continuation account, while the auxiliary probes rule against treating the effect as only a final-token, uncertainty, or one-step continuation artifact.

\begin{table*}[t]
\centering
\scriptsize
\setlength{\tabcolsep}{3pt}
\renewcommand{\arraystretch}{1.08}
\begin{tabular}{p{2.5cm}p{3.1cm}p{5.4cm}p{4.1cm}}
\toprule
Probe family & Evidence tier & Paper-facing result & Claim boundary \\
\midrule
Qwen forced-prefix & Retained & Anti has higher continuation than baseline overall (+1.54 chars, CI [-0.01, 3.66]) and a clearer \texttt{scope\_minimal\_sufficient} slice (+3.22, CI [0.56, 7.28]); immediate-stop counts are 62/100, 48/100, and 63/100 for baseline/anti/minimal. & Supports late-continuation pressure, but the main-family continuation gap is modest. \\
Qwen planning/stopping annotation & Retained & Overall annotation share is 10.0\% planning, 23.75\% stopping, and 10.0\% mixed; \texttt{scope\_minimal\_sufficient} is more planning-heavy, while \texttt{do\_not\_add\_unasked\_help} is more stopping-heavy. & Gives content-level interpretation of excess text; not a circuit-level mechanism. \\
Bundled-control probes & Retained & Anti remains elevated across nearby pullback prompts: asymmetric-control negative mean 71.22 vs.\ 25.89 baseline; bundled-generalization negative mean 59.00 vs.\ 26.42 baseline. & Supports bundled assistant-policy behavior rather than one isolated length knob. \\
EOS/tailspan, uncertainty, branchpoint & Retained for narrowing & EOS helps less than tail-span supervision; long anti outputs are not explained by higher uncertainty; branchpoint signals are too weak to explain the full result. & Used to disfavor shallow single-factor reductions, not as a standalone causal estimate. \\
Compression/pruning & Retained for Qwen, echoed in SmolLM2/7B & Same-information compression leaves anti outputs relatively long, while true pruning shortens more; SmolLM2 repeats the direction (baseline 278.0$\rightarrow$172.9$\rightarrow$99.8, anti 337.1$\rightarrow$251.7$\rightarrow$117.5). & Supports content-budgeting pressure; 7B compression/pruning is treated as auxiliary because it is prompt-sensitive. \\
\bottomrule
\end{tabular}
\caption{Mechanism-probe summary and claim scope. The table separates behavioral evidence from auxiliary mechanism-narrowing probes and supports the mixed planning/stopping account.}
\label{tab:mechanism-evidence-boundary}
\end{table*}

\noindent Table~\ref{tab:mechanism-numeric-audit} gives the compact numerical audit behind the qualitative mechanism-narrowing entries in Table~\ref{tab:mechanism-evidence-boundary}. These numbers make explicit why shallow single-factor accounts are insufficient and why the mixed planning/stopping interpretation is the best behavioral summary of the probes.

\begin{table*}[t]
\centering
\scriptsize
\setlength{\tabcolsep}{3pt}
\renewcommand{\arraystretch}{1.08}
\begin{tabular}{p{2.4cm}p{5.2cm}p{7.4cm}}
\toprule
Probe & Numerical summary & Reading \\
\midrule
EOS/tailspan & Mean length over 48 user-boundary rows: prefix50 150.54, true-EOS 116.00, tailspan 105.02. Boundary-mode means for avoid-UA / scope-min / no-help are 147.33 / 152.67 / 152.42, 116.50 / 110.00 / 110.00, and 115.17 / 102.83 / 104.92. & True EOS reduces length, while tailspan supervision reduces it further, so the probe does not support a pure final-token EOS account. \\
Uncertainty & Median split over 24 anti uncertainty rows: short outputs average 27.75 chars, entropy 3.398, margin 0.211; long outputs average 103.75 chars, entropy 2.550, margin 0.301. Mode means are neutral 56.33 chars / entropy 2.926 / margin 0.252 and minimal-sufficient 75.17 chars / entropy 3.021 / margin 0.260. & Longer outputs are not higher-entropy or lower-margin in the simple uncertainty-compensation pattern. \\
Branchpoint & Across 12 branchpoint probes, the average-logprob winner is \texttt{stop} in 12/12 cases; mean continue-minus-stop average-logprob gap is -1.944, with 0/12 positive continue gaps. & Local branchpoint scoring does not support reducing the main result to a simple one-step continue preference. \\
\bottomrule
\end{tabular}
\caption{Compact numerical audit for mechanism-narrowing probes. EOS/tailspan values are mean Chinese-character lengths; uncertainty entries report generation-time entropy and top1--top2 margin; branchpoint entries report average-logprob comparisons for candidate continuations.}
\label{tab:mechanism-numeric-audit}
\end{table*}

\section{Planning-vs.-Stopping Raw Summary}
\label{app:planning-stopping}

\noindent Table~\ref{tab:planning-stopping-raw} reports the raw percentages behind Figure~\ref{fig:planning-stopping}, so the main-text decomposition can be checked directly against the underlying counts.

\begin{table}[H]
\centering
\footnotesize
\resizebox{\columnwidth}{!}{%
\begin{tabular}{lcccc}
\toprule
Slice & planning & stopping & mixed & note \\
\midrule
overall & 10.0\% & 23.75\% & 10.0\% & plan $=.338$, stop $=.662$ \\
\texttt{scope\_minimal\_sufficient} & 17.5\% & 12.5\% & --- & planning-heavy \\
\texttt{do\_not\_add\_unasked\_help} & 2.5\% & 35.0\% & --- & stopping-heavy \\
\bottomrule
\end{tabular}
}
\caption{Raw planning-vs.-stopping percentages corresponding to Figure~\ref{fig:planning-stopping}. The overall row covers 80 annotated Qwen responses; plan/stop notes report mean planning and stopping scores.}
\label{tab:planning-stopping-raw}
\end{table}

\section{Qwen Forced-prefix Breakdown}
\label{app:qwen-fp-breakdown}

\noindent Table~\ref{tab:qwen-fp-breakdown} reports the forced-prefix continuation slices behind the FP column in Table~\ref{tab:main-family-results}. The overall gap is directionally positive but modest, while the \texttt{scope\_minimal\_sufficient} slice shows the clearest late-continuation gap.

\begin{table}[H]
\centering
\footnotesize
\resizebox{\columnwidth}{!}{%
\begin{tabular}{lcc}
\toprule
Slice & $\Delta$ anti-base & 95\% CI \\
\midrule
overall & +1.54 & [-0.01, 3.66] \\
neutral & -0.14 & [-1.48, 0.84] \\
\texttt{scope\_minimal\_sufficient} & +3.22 & [0.56, 7.28] \\
\bottomrule
\end{tabular}
}
\caption{Forced-prefix continuation gaps in the Qwen main family. Entries are anti-minus-baseline deltas in mean continuation characters.}
\label{tab:qwen-fp-breakdown}
\end{table}

\begin{table}[H]
\centering
\footnotesize
\resizebox{\columnwidth}{!}{%
\begin{tabular}{lccc}
\toprule
Family & Immediate stops & Total & Stop rate \\
\midrule
\texttt{baseline} & 62 & 100 & 62\% \\
\texttt{anti\_underanswer} & 48 & 100 & 48\% \\
\texttt{minimal\_boundary} & 63 & 100 & 63\% \\
\bottomrule
\end{tabular}
}
\caption{Immediate-stop counts in the Qwen forced-prefix probe. Immediate stop means zero-character continuation after the minimum-sufficient prefix.}
\label{tab:qwen-fp-stop-counts}
\end{table}

\section{Asymmetric-controllability and Bundled-generalization Raw Means}
\label{app:asym-bundle-raw}

\noindent Table~\ref{tab:asym-bundle-raw} gives the raw paired means behind the asymmetric-controllability and bundled-generalization discussion in Section~\ref{sec:mechanism}. It separates the pullback side from the neighboring positive side so the bundled reading can be checked directly against the probe outputs.

\begin{table*}[t]
\centering
\scriptsize
\resizebox{\textwidth}{!}{%
\begin{tabular}{p{2.3cm}p{2.0cm}cccccc}
\toprule
Probe & Axis & base pullback & base positive & anti pullback & anti positive & min pullback & min positive \\
\midrule
Asym. ctrl & scope & 26.00 & 23.00 & 87.17 & 74.50 & 28.17 & 35.17 \\
Asym. ctrl & interaction & 25.83 & 23.17 & 76.67 & 28.83 & 33.50 & 35.33 \\
Asym. ctrl & help & 25.83 & 27.00 & 49.83 & 68.67 & 43.33 & 49.50 \\
\midrule
Bundled gen. & scope & 33.17 & 23.00 & 55.50 & 51.17 & 26.17 & 38.83 \\
Bundled gen. & interaction & 24.50 & 22.00 & 56.67 & 64.17 & 25.67 & 28.33 \\
Bundled gen. & help & 24.00 & 23.33 & 58.67 & 52.17 & 24.67 & 33.83 \\
Bundled gen. & caution & 24.00 & 25.50 & 65.17 & 43.67 & 23.50 & 25.33 \\
\bottomrule
\end{tabular}
}
\caption{Raw paired mean Chinese-character lengths for asymmetric controllability and bundled generalization in the Qwen main family. Pullback columns correspond to boundary-narrowing prompts; positive columns correspond to nearby expansion or addition prompts.}
\label{tab:asym-bundle-raw}
\end{table*}

\section{Second-family Mechanism Echoes}
\label{app:smollm2-mechanism}

\noindent This section gives the SmolLM2 checks summarized in Section~\ref{sec:external}. These checks serve as second-family mechanism echoes; the full mechanism suite remains concentrated in the Qwen family.

\begin{table*}[t]
\centering
\scriptsize
\setlength{\tabcolsep}{2pt}
\renewcommand{\arraystretch}{1.08}
\begin{tabularx}{\textwidth}{p{2.35cm}XXX}
\toprule
Check & Baseline & Anti & Reading \\
\midrule
Forced-prefix continuation 
& Mean continuation 72.7; scope-only mean 42.1 
& Mean continuation 131.5; scope-only mean 80.6 
& Anti continues more after a minimum-sufficient prefix. \\

Compression/pruning 
& 278.0 $\rightarrow$ 172.9 $\rightarrow$ 99.8 
& 337.1 $\rightarrow$ 251.7 $\rightarrow$ 117.5 
& Same-information compression leaves anti longer; true pruning shortens both more strongly. \\

Planning/stopping annotation 
& Mean 0.375 / 0.417; counts: 14 none, 4 stop, 5 plan, 0 mixed, 1 unclear
& Mean 0.625 / 1.042; counts: 7 none, 9 stop, 2 plan, 6 mixed, 0 unclear
& Anti is heavier on both planning and stopping dimensions. \\
\bottomrule
\end{tabularx}
\caption{SmolLM2 mechanism checks. Compression/pruning reports original $\rightarrow$ same-information compression $\rightarrow$ true pruning. Planning/stopping means are reported as planning/stopping scores; counts are per-model dominant-label counts over 24 labeled rows.}
\label{tab:smollm2-mechanism-checks}
\end{table*}

\section{Shared-system and 7B-scale Robustness}
\label{app:shared-system-7b}

\noindent Table~\ref{tab:shared-system-main} reports the shared-system boundary-control checks. These checks use a common system prompt across policy families, so the anti/baseline/minimal distinction is carried by the supervised response behavior.

\begin{table}[H]
\centering
\footnotesize
\resizebox{\columnwidth}{!}{%
\begin{tabular}{lccc}
\toprule
Check & \texttt{baseline} & \texttt{anti} & \texttt{minimal} \\
\midrule
Qwen1.5B shared-system mean & 49.75 & 93.30 & 31.55 \\
Qwen1.5B shared-system median & 44.00 & 93.50 & 26.00 \\
Qwen7B shared-system mean & 33.05 & 81.67 & 27.34 \\
\bottomrule
\end{tabular}
}
\caption{Shared-system boundary-control summaries. Means average boundary-control evaluation modes under a common system prompt.}
\label{tab:shared-system-main}
\end{table}

The paired anti-minus-baseline boundary-control deltas are +43.55 characters for Qwen1.5B (95\% CI [37.35, 49.61]) and +48.62 characters for Qwen7B (95\% CI [44.25, 52.99]), computed over the 150 aligned case-mode rows in each shared-system line.

\noindent Table~\ref{tab:7b-mechanism-traces-app} summarizes selected 7B mechanism traces. The same direct ordering appears at the larger scale, with \texttt{anti\_underanswer} much less likely to stop after a minimum-sufficient prefix and still elevated under bundled controls.

\begin{table}[H]
\centering
\footnotesize
\begin{tabular}{lccc}
\toprule
7B probe & \texttt{baseline} & \texttt{anti} & \texttt{minimal} \\
\midrule
FP mean & 11.40 & 52.90 & 6.89 \\
Immediate stops & 49/100 & 3/100 & 71/100 \\
Bundled gen. overall & 90.10 & 129.42 & 119.06 \\
\bottomrule
\end{tabular}
\caption{7B selected mechanism traces. Immediate stops use the probe's annotated \texttt{stop\_immediate} continuation class rather than raw zero-character length.}
\label{tab:7b-mechanism-traces-app}
\end{table}

For the 7B forced-prefix probe, the paired anti-minus-baseline continuation delta is +41.50 characters over 100 aligned rows (95\% CI [36.62, 46.39]).

\noindent Table~\ref{tab:7b-asym-bundle} reports additional 7B probe details. The asymmetric-control result is axis-sensitive: \texttt{minimal\_boundary} is narrowest on the explicit \texttt{ask\_minimal} mode, while the overall asymmetric-control mean is affected by non-scope axes.

\begin{table}[H]
\centering
\footnotesize
\begin{tabular}{lccc}
\toprule
Probe & \texttt{baseline} & \texttt{anti} & \texttt{minimal} \\
\midrule
Asym. ctrl overall & 87.14 & 126.03 & 129.56 \\
\texttt{ask\_minimal} & 56.83 & 128.00 & 48.17 \\
\bottomrule
\end{tabular}
\caption{7B asymmetric-control summary. Bundled-generalization overall means are reported in Table~\ref{tab:7b-mechanism-traces-app}.}
\label{tab:7b-asym-bundle}
\end{table}

\noindent Table~\ref{tab:7b-compression-pruning} gives the 7B compression-vs.-pruning summary. We use this as auxiliary mechanism evidence because the compression/pruning split is more prompt- and family-sensitive than the boundary-control and forced-prefix checks.

\begin{table}[H]
\centering
\footnotesize
\begin{tabular}{lccc}
\toprule
Family & neutral & compression & pruning \\
\midrule
\texttt{baseline} & 32.62 & 27.75 & 31.25 \\
\texttt{anti\_underanswer} & 100.62 & 79.25 & 83.12 \\
\texttt{minimal\_boundary} & 26.00 & 22.75 & 19.38 \\
\bottomrule
\end{tabular}
\caption{7B compression-vs.-pruning summary. Entries are mean response lengths; compression refers to same-information compression. This auxiliary probe is not assumed to be monotonic across prompt variants.}
\label{tab:7b-compression-pruning}
\end{table}

\section{External Cluster Raw Counts}
\label{app:external-cluster-raw}

\noindent Table~\ref{tab:external-cluster-raw} gives the raw negative-mode annotation counts behind Figure~\ref{fig:external-cluster}. It lets the main-text cluster summary be traced back to its original \texttt{obey/partial/disobey} counts.

\begin{table}[H]
\centering
\scriptsize
\setlength{\tabcolsep}{3pt}
\begin{tabular}{lccc}
\toprule
Model & \texttt{caveat} & \texttt{next\_step} & \texttt{moralizing} \\
\midrule
\texttt{qwen} & 7 / 4 / 1 & 13 / 3 / 0 & 3 / 5 / 0 \\
\texttt{ds\_v4\_flash} & 8 / 1 / 3 & 7 / 9 / 0 & 4 / 4 / 0 \\
\texttt{gpt5mini\_v4} & 4 / 6 / 2 & 10 / 5 / 0 & 2 / 5 / 0 \\
\texttt{ds7b\_chat\_v4} & 4 / 1 / 7 & 4 / 5 / 7 & 0 / 6 / 2 \\
\texttt{gpt5\_1\_v4} & 10 / 0 / 2 & 14 / 2 / 0 & 4 / 4 / 0 \\
\texttt{claude\_s46\_v4} & 9 / 0 / 3 & 13 / 3 / 0 & 6 / 2 / 0 \\
\bottomrule
\end{tabular}
\caption{Raw negative-mode counts for the external cluster benchmark. Entries are \texttt{obey / partial / disobey}; row totals differ by axis, so Figure~\ref{fig:external-cluster} should be read as a raw audit heatmap rather than a normalized prevalence chart.}
\label{tab:external-cluster-raw}
\end{table}

\section{Exploratory Discrimination Check}
\label{app:exploratory-discrimination}

\noindent We ran a small prompt-side discrimination check on stock \texttt{SmolLM2-1.7B-Instruct} to compare boundary pullback with formatting-only and explanation-only pullback. This exploratory check provides boundary-setting evidence only.

\begin{table}[H]
\centering
\footnotesize
\begin{tabular}{lll}
\toprule
Pullback type & Negative mode & Mean length \\
\midrule
Boundary & \texttt{no\_extra\_help} & 353.8 \\
Explanation & \texttt{answer\_only} & 122.2 \\
Formatting & \texttt{plain\_sentence} & 133.2 \\
\bottomrule
\end{tabular}
\caption{Exploratory discrimination check on pullback types. Lower mean length indicates easier realization of the negative/pullback instruction.}
\label{tab:exploratory-discrimination}
\end{table}

\section{Partial Controlled-family Extension for \texttt{caveat}}
\label{app:caveat-extension}

\noindent Because \texttt{caveat/completeness} was the most stable external prompt-side axis and the clearest controlled-extension candidate, we tested whether it could be partially transferred into a controlled-family setting. This line provides partial evidence that this external prompt-side axis has a controlled-training analogue, while falling short of a second clean family.

\begin{table}[H]
\centering
\footnotesize
\begin{tabular}{lccc}
\toprule
Evaluation mode & obey & partial & disobey \\
\midrule
\texttt{ask\_no\_extra\_caveat} & 9 & 3 & 0 \\
\texttt{ask\_more\_complete\_caveat} & 7 & 2 & 3 \\
\bottomrule
\end{tabular}
\caption{Partial controlled-family extension for \texttt{caveat/completeness}. Counts report \texttt{obey/partial/disobey} outcomes under the two evaluation directions.}
\label{tab:caveat-extension}
\end{table}

\section{Evidence Tiers}
\label{app:evidence-tiers}

\noindent This appendix records the evidence tiers used throughout the paper, separating retained claim-bearing evidence from partial, exploratory, and excluded lines.

\begin{table*}[t]
\centering
\scriptsize
\setlength{\tabcolsep}{3pt}
\renewcommand{\arraystretch}{1.08}
\begin{tabular}{p{1.8cm}p{6.0cm}p{8.0cm}}
\toprule
Tier & Representative lines & Role \\
\midrule
Retained & Main \texttt{baseline/anti/minimal}; EOS/tailspan, uncertainty, compression/pruning, forced-prefix, asymmetric/bundled, planning/stopping; \texttt{baseline\_then\_anti}; \texttt{anti$\rightarrow$pref\_minimal}; SmolLM2 direct replication; shared-system Qwen1.5B; 7B boundary-control/FP; external cluster & Carries the main phenomenon, mechanism narrowing, stage amplification/reversal, and bounded-generality claims \\
Partial & \texttt{baseline\_then\_preference\_expand}; \texttt{minimal$\rightarrow$pref\_baseline}; \texttt{more\_complete\_caveat}; SmolLM2 mechanism echoes; 7B asymmetric/bundled/compression summaries & Supports weaker preference-like amplification, midpoint pullback, partial caveat extension, and auxiliary mechanism checks \\
Exploratory & affect and stance axes; additional high-level contrasts; older minimal-side diagnostics; \texttt{caveat} v1/v3; SmolLM2 boundary-vs-format/explanation check & Boundary-setting and negative exploration only \\
Excluded & old one-layer lines; broken minimal-side pair constructions; confounded outward-from-minimal runs; known data/anchor/loader confounds & Excluded from main conclusions \\
\bottomrule
\end{tabular}
\caption{Evidence tiers used in the paper.}
\label{tab:evidence-tiering}
\end{table*}

\section{Stage-line Filtering and Confound Checks}
\label{app:stage-line-filtering}

\noindent Table~\ref{tab:stage-line-filtering} summarizes filtering and reconstruction checks applied before retaining the stage and reversal lines. These checks are included for auditability; excluded or confounded runs stay outside the paper's conclusions.

\begin{table*}[t]
\centering
\scriptsize
\setlength{\tabcolsep}{3pt}
\renewcommand{\arraystretch}{1.08}
\begin{tabular}{p{3.4cm}p{5.9cm}p{6.2cm}}
\toprule
Potential confound & Filtering / reconstruction check & Why it matters \\
\midrule
Non-baseline pair-anchor mismatch & Rewrote anchors to be policy-consistent on non-baseline sides before retaining reversal lines. & Reduces risk that apparent pullback comes from prompt-anchor mismatch. \\
Broken or over-jumping minimal-side pairs & Filtered large target jumps, removed broken one-layer constructions, and retained only the cleaner midpoint line. & Separates weak midpoint recovery from broken pair construction or overly large shifts. \\
Preference-stage data-loading instability & Rebuilt pairwise data with local JSONL loading, tokenization, and fixed anchors. & Reduces preprocessing artifacts as an explanation for retained preference-like effects. \\
\bottomrule
\end{tabular}
\caption{Filtering and confound checks for retained stage lines.}
\label{tab:stage-line-filtering}
\end{table*}

\section{Stage Delta Details}
\label{app:stage-delta-uncertainty}

\noindent Table~\ref{tab:stage-delta-uncertainty} reports bootstrap intervals for the stage deltas in Table~\ref{tab:stage-and-reversal}. The stage rows use a 12-case stage set, while the direct-family references use the larger direct-family evaluation rows. Because the case pools are not identical, these intervals are independent-bootstrap intervals over the stage row and its direct-family source row, not paired-bootstrap intervals.

\begin{table*}[t]
\centering
\scriptsize
\setlength{\tabcolsep}{3pt}
\renewcommand{\arraystretch}{1.08}
\begin{tabular}{lccc}
\toprule
Line / comparison & avoid-UA $\Delta$ [95\% CI] & scope-min $\Delta$ [95\% CI] & FP $\Delta$ [95\% CI] \\
\midrule
\texttt{baseline\_then\_anti} vs.\ direct baseline & +68.64 [52.17, 85.16] & +75.26 [55.12, 96.94] & +53.29 [45.63, 60.59] \\
\texttt{baseline\_then\_pref\_expand} vs.\ direct baseline & +27.64 [-0.30, 61.22] & +25.01 [4.14, 48.69] & +25.46 [9.78, 42.09] \\
\texttt{anti$\rightarrow$pref\_minimal} vs.\ direct anti & -19.53 [-30.06, -10.44] & -27.38 [-41.36, -15.94] & -3.08 [-5.73, -0.82] \\
\texttt{minimal$\rightarrow$pref\_baseline} vs.\ direct minimal & +5.84 [-13.94, 33.78] & -5.61 [-16.43, 4.80] & +5.91 [-1.78, 18.89] \\
\bottomrule
\end{tabular}
\caption{Independent-bootstrap uncertainty for stage deltas. Boundary-control stage rows have 12 examples per mode; direct-family boundary-control references have 50 examples per mode. Forced-prefix stage rows have 12 examples; direct-family forced-prefix references have 100 examples.}
\label{tab:stage-delta-uncertainty}
\end{table*}

\section{Negative-control and Non-retained Contrast Axes}
\label{app:negative-controls}

\noindent The main external-cluster figure reports axes that showed residual pullback failure. Table~\ref{tab:negative-controls} adds contrastive and non-retained lines used to delimit hard pullback from generic high-level steering. These lines are boundary-setting evidence rather than full negative results.

\begin{table*}[t]
\centering
\scriptsize
\setlength{\tabcolsep}{3pt}
\renewcommand{\arraystretch}{1.08}
\begin{tabular}{p{3.0cm}p{6.2cm}p{6.3cm}}
\toprule
Axis & Retained evidence & Boundary reading \\
\midrule
\texttt{emoji\_only} & Phase-2 boundary-control means stay below the anti-underanswer regime; unsolicited emoji appears in \(0/48\) rows overall and \(0/36\) boundary-control rows. & Style marker is inducible but fades under boundary evaluation; same-framework contrast case. \\
\texttt{affect\_attuned} vs.\ \texttt{affect\_flat} & Runnable stage-2 attempts, but no stable final obedience-count table retained. & Evaluation drifted into comfort/support/length bundles, leaving no clean suppressibility axis. \\
\texttt{deferential\_uncertain} vs.\ \texttt{decisive\_direct} & Stage-2 attempts and external v5 diagnostic; no retained final count table. & Prompting moved the axis, while training did not stabilize a clean sticky family. \\
\texttt{clarify\_first} vs.\ \texttt{answer\_first} & Qwen phase-2 boundary-control means remain low, with explicit question-asking in \(0/36\) boundary-control rows. & No overt clarify-first reversion under boundary-control modes; contrastive boundary case. \\
\bottomrule
\end{tabular}
\caption{Negative-control and non-retained contrast axes used to delimit the external cluster. These lines provide boundary-setting evidence for the external cluster.}
\label{tab:negative-controls}
\end{table*}

\section{Non-retained Extra Directions}
\label{app:non-retained-directions}

\noindent We also explored several additional high-level directions that did not become retained axes. \texttt{affect\_attuned vs.\ affect\_flat} produced runnable and interpretable outputs, but full evaluation drifted into comfort/support/length bundles. \texttt{deferential\_uncertain vs.\ decisive\_direct} was already strongly moved by the prompt alone, and stage-2 training did not turn it into a clean sticky family. Older minimal-side outward lines are not used because they were entangled with broken pair construction, over-large target jumps, or implementation confounds.

These lines document the search boundary. They suggest that sticky pullback failures are selective across high-level assistant traits, while the main claims rely on the retained evidence above.

\end{document}